\documentclass{article}

\begin{document}

\title{Evolved Preambles for MAX-SAT Heuristics}

\author{Lu\'\i s O. Rigo Jr.\\
Valmir C. Barbosa\thanks{Corresponding author (valmir@cos.ufrj.br).}\\
\\
Programa de Engenharia de Sistemas e Computa\c c\~ao, COPPE\\
Universidade Federal do Rio de Janeiro\\
Caixa Postal 68511\\
21941-972 Rio de Janeiro - RJ, Brazil}

\date{}

\maketitle

\begin{abstract}
MAX-SAT heuristics normally operate from random initial truth assignments to the
variables. We consider the use of what we call preambles, which are sequences of
variables with corresponding single-variable assignment actions intended to be
used to determine a more suitable initial truth assignment for a given problem
instance and a given heuristic. For a number of well established MAX-SAT
heuristics and benchmark instances, we demonstrate that preambles can be evolved
by a genetic algorithm such that the heuristics are outperformed in a
significant fraction of the cases.

\end{abstract}

\section{Introduction}

Given a set $V$ of Boolean variables and a set of disjunctive clauses on
literals from $V$ (i.e., variables or their negations), MAX-SAT asks for a truth
assignment to the variables that maximizes the number of clauses that are
satisfied (i.e., made true by that assignment). MAX-SAT is NP-hard \cite{gj79}
but can be approximated in polynomial time, though not as close to the optimum
as one wishes (cf.\ \cite{ausiello99} and references therein, as well as
\cite{dantsin01} for the restricted case of three-literal clauses). MAX-SAT has
enjoyed a paradigmatic status over the years, not only because of its close
relation to SAT, the first decision problem to be proven NP-complete, but also
because of its importance to other areas (e.g., constraint satisfaction in
artificial intelligence \cite{d03}).

Since NP-hardness is a property of worst-case scenarios, the difficulty of
actually solving a specific instance of an NP-hard problem varies widely with
both the instance's size and internal structure. In fact, in recent years it has
become increasingly clear that small changes in either can lead to significant
variation in an algorithm's performance, possibly even to a divide between the
instance's being solvable or unsolvable by that algorithm given the
computational resources at hand and the time one is willing to spend
\cite{hw05}. Following some early groundwork \cite{rice76}, several attempts
have been made at providing theoretical foundations or practical guidelines for
automatically selecting which method to use given the instance
\cite{russell95,minton96,fink98,gomes01,lagoudakis01,leyton-brown03,vassilevska06}.

Here we investigate a different, though related, approach to method selection in
the case of MAX-SAT instances. Since all MAX-SAT heuristics require an initial
truth assignment to the variables, and considering that this is invariably
chosen at random, a natural question seems to be whether it is worth spending
some additional effort to determine an initial assignment that is better suited
to the instance at hand. Once we adopt this two-stage template comprising an
initial-assignment selection in tandem with a heuristic, fixing the latter
reduces the issue of method selection to that of identifying a procedure to
determine an appropriate initial assignment. We refer to this procedure as a
preamble to the heuristic. As we demonstrate in the sequel, for several
state-of-the-art heuristics and problem instances the effort to come up with an
appropriate preamble pays off in terms of better solutions for the same amount
of time.

We proceed in the following manner. First, in Section~\ref{preambles}, we define
what preambles are in the case of MAX-SAT. Then we introduce an evolutionary
method for preamble determination in Section~\ref{methods} and give
computational results in Section~\ref{results}. We close with concluding remarks
in Section~\ref{conclusions}.

\section{MAX-SAT preambles}
\label{preambles}

Let $n$ be the number of variables in $V$. Given a MAX-SAT instance on $V$, a
preamble $p$ of length $\ell$ is a sequence of pairs
$\langle v_1,a_1\rangle,\langle v_2,a_2\rangle,\ldots,\langle v_\ell,a_\ell\rangle$,
each representing a computational step to be taken as the preamble is played
out. In this sequence, and for $1\le k\le\ell$, the $k$th pair is such that
$v_k\in V$ and $a_k$ is one of $2$, $1$, or $0$, indicating respectively whether
to leave the value of $v_k$ unchanged, to act greedily when choosing a value for
$v_k$, or to act contrarily to such greedy assignment. A preamble need not
include all $n$ variables, and likewise a variable may appear more than once in
it.

The greedy action to which $a_k$ sometimes refers assigns to $v_k$ the truth
value that maximizes the number of satisfied clauses at that point in the
preamble. Algorithmically, then, playing out $p$ is equivalent to performing the
following steps from some initial truth assignment to the variables in $V$:
\begin{enumerate}
\item $k:=1$.
\item If $a_k=2$, then proceed to Step 5.
\item Given the current values of all other variables, compute the number of
clauses that get satisfied for each of the two possible assignments to $v_k$.
\item If $a_k=1$, then set $v_k$ to the truth value yielding the greatest number
of satisfied clauses. If $a_k=0$, then do the opposite. Break ties randomly.
\item $k:=k+1$. If $k\le\ell$, then proceed to Step 2.
\end{enumerate}
We use random initial assignments exclusively. A MAX-SAT preamble, therefore,
can be thought of as isolating such initial randomness from the heuristic proper
that is to follow the preamble. Instead of starting the heuristic at its usual
random initial assignment, we start it at the assignment determined by running
the preamble.

It is curious to note that, as defined, a preamble generalizes the sequence of
steps generated by the simulated annealing method \cite{kgv83} when applied to
MAX-SAT. In fact, what simulated annealing does in this case, following one of
its variations \cite{gg84,b93}, is to choose $v_k$ by cycling through the
members of $V$ and then let $a_k$ be either $1$ or $0$ with the Boltzmann-Gibbs
probability. At the high initial temperatures the two outcomes are nearly
equally probable, but the near-zero final temperatures imply $a_k=1$ (i.e., be
greedy) with high probability. The generalization that comes with our definition
allows for various possibilities of preamble construction, as the evolutionary
procedure we describe next.

\section{Methods}
\label{methods}

Given a MAX-SAT instance and heuristic $H$, our approach is to evolve the best
possible preamble to $H$. We do so through a genetic algorithm of the
generational type \cite{m96}. The description that follows refers to parameter
values that were determined in an initial calibration phase. This phase used the
heuristics
\textit{gsat} \cite{selman92},
\textit{gwsat} \cite{selman93},
\textit{hsat} \cite{gent93},
\textit{hwsat} \cite{gent95},
\textit{gsat-tabu} \cite{mazure97},
\textit{novelty} \cite{mcallester97},
\textit{walksat-tabu} \cite{mcallester97},
\textit{adaptnovelty+} \cite{hoos02},
\textit{saps} \cite{hutter02}, and
\textit{sapsnr} \cite{tompkins04}
as heuristic $H$, and also the instances
\texttt{C880mul},
\texttt{am\_8\_8},
\texttt{c3540mul},
\texttt{term1mul}, and
\texttt{vdamul}
from \cite{sat04}. Each of the latter involves variables that number in the
order of $10^4$ and clauses numbering in the order of $10^5$. Moreover, not all
optima are known (cf.\ Section~\ref{results}).

The genetic algorithm operates on a population of $50$ individuals, each being a
preamble to heuristic $H$. The fitness of individual $p$ is computed as follows.
First $p$ is run from $10$ random truth assignments to the variables, then $H$
is run from the truth assignment resulting from the run of $p$ that satisfied
the most clauses (ties between runs of $p$ are broken randomly). Let $R(p)$
denote the number of clauses satisfied by this best run of $p$ and $S_H(p)$ the
number of clauses satisfied after $H$ is run. The fitness of individual $p$ is
the pair $\langle S_H(p),R(p)\rangle$. Whenever two individuals' fitnesses are
compared, ties are first broken lexicographically, then randomly. Selection is
always performed from linearly normalized fitnesses, the fittest individual of
the population being $20$ times as fit as the least fit.

For each MAX-SAT instance and each heuristic $H$, we let the genetic algorithm
run for a fixed amount of time, during which a new population is repeatedly
produced from the current one and replaces it. The initial population comprises
individuals of maximum length $1.5n$, each created randomly to contain at least
$0.4n$ distinct variables. The process of creating each new population starts by
an elitist step that transfers the $20\%$ fittest individuals from the current
population to the new one. It then repeats the following until the new
population is full.

First a decision is made as to whether crossover (with probability $0.25$) or
mutation (with probability $0.75$) is to be performed. For crossover two
individuals are selected from the current population and each is partitioned
into three sections for application of the standard two-point crossover
operator. The partitioning is done randomly, provided the middle section
contains exactly $0.4n$ distinct variables, which is always possible by
construction of the initial population (though at times either of the two
extreme sections may turn out to be empty). The resulting two individuals (whose
lengths are no longer bounded by $1.5n$) are added to the new population. For
mutation a single individual is selected from the current population and
$50\%$ of its pairs are chosen at random. Each of these, say the $k$th pair,
undergoes either a random change to both $v_k$ and $a_k$ (if this will leave the
individual with at least $0.4n$ distinct variables) or simply a random change to
$a_k$ (otherwise). The mutant is then added to the new population.

The calibration phase referred to above also yielded three champion heuristics,
viz.\ \textit{novelty}, \textit{walksat-tabu}, and \textit{adaptnovelty+}. The
results we give in Section~\ref{results} refer exclusively to these, used either
in conjunction with the genetic algorithm as described above or by themselves.
In the latter case each heuristic is run repeatedly, each time from a new random
truth assignment to the variables, until the same fixed amount of time used for
the genetic algorithm has elapsed. The result reported by the genetic algorithm
refers to the fittest individual in the last population to have been filled
during that time. As for the heuristic, in order to compare its performance with
that of the genetic algorithm as fairly as possible the result that is reported
is the best one obtained after every $50$ repetitions.

All experiments were performed from within the \textit{UBCSAT} environment
\cite{tompkins04pkg}. Optima, whenever possible, were discovered separately via
the 2010 release of the \textit{MSUnCore} code to solve MAX-SAT exactly
\cite{msu09}.

\section{Computational results}
\label{results}

In our experiments we tackled all $100$ instances of the 2004 SAT competition
\cite{sat04}, henceforth referred to as the 2004 dataset, and all $112$
instances of the 2008 MAX-SAT evaluation \cite{maxsat08}, henceforth referred to
as the 2008 dataset. The time allotted for each instance to the genetic
algorithm or each of the three heuristics by itself was of $60$ minutes, always
on identical hardware and software, always with exclusive access to the system.
We report exclusively on the hardest instances from either dataset, here defined
to be those for which \textit{MSUnCore} found no answer as a result of being
stymied by the available 4 gigabytes of RAM and the system's inability to
perform further swapping. There are $51$ such instances in the 2004 dataset,
$11$ in the 2008 dataset, totaling $62$ instances.

Our results are given in Tables~\ref{nov} through~\ref{anov}, respectively for
$H=\textit{novelty}$, $H=\textit{walksat-tabu}$, and $H=\textit{adaptnovelty+}$.
Each table contains a row for each of the $62$ instances, with a horizontal line
separating the instances of the 2004 dataset from those of the 2008 dataset.
For each instance the number $n$ of variables is given, as well as the number of
clauses ($m$) and results for the genetic algorithm and for the heuristic in
question by itself. These results are the number of satisfied clauses and the
time at which this solution was first found during the allotted $60$ minutes.
Missing results indicate either that no population could be filled during this
time (in the case of the genetic algorithm) or that no batch of $50$ runs of the
heuristic could be finished.

\begin{table}[p]
\centering
\scriptsize
\caption{Results for $H=\textit{novelty}$. Times are given in minutes.}
\begin{tabular}{lcccrcr}
\hline & & & \multicolumn{2}{c}{Genetic algorithm} & \multicolumn{2}{c}{\textit{novelty} alone} \\
Instance & $n$ & $m$ & {Num.\ sat.\ cl.} & Time & {Num.\ sat.\ cl.} & Time \\ \hline
\texttt{c3540mul} &5248 & 33199 & 33176 & 17.965 & 33180 & 6.291 \\
\texttt{c6288mul} &9540 & 61421 & 61375 & 21.391 & 61375 & 6.706 \\
\texttt{dalumul} &9426 & 59991 & 59972 & 59.961 & 59973 & 5.712 \\
\texttt{frg1mul} & 3230 & 20575 & \textbf{20574} & \textbf{0.543} & 20574 & 1.390 \\
\texttt{k2mul} & 11680 & 74581 & \textbf{74524} & 1.729 & 74516 & 23.493 \\
\texttt{x1mul} & 8760 & 55571 & \textbf{55570} & \textbf{1.550} & 55570 & 5.383 \\
\texttt{am\_6\_6} & 2269 & 7814 & \textbf{7813} & \textbf{0.263} & 7813 & 0.267 \\
\texttt{am\_7\_7} & 4264 & 14751 & \textbf{14744} & 55.270 & 14733 & 50.688 \\
\texttt{am\_8\_8} & 7361 & 25538 & 25331 & 37.914 & 25332 & 36.184 \\
\texttt{am\_9\_9} & 11908 & 41393 & \textbf{40982} & 1.827 & 40970 & 36.322 \\
\texttt{li-exam-61} & 28147 & 108436 & 108011 & 47.997 & 108041 & 22.993 \\
\texttt{li-exam-62} & 28147 & 108436 & 107999 & 56.149 & 108006 & 27.760 \\
\texttt{li-exam-63} & 28147 & 108436 & 107998 & 29.129 & 108003 & 19.108 \\
\texttt{li-exam-64} & 28147 & 108436 & 107987 & 58.590 & 108013 & 34.106 \\
\texttt{li-test4-100} & 36809 & 142491 & 141844 & 34.560 & 141856 & 35.189 \\
\texttt{li-test4-101} & 36809 & 142491 & 141858 & 40.733 & 141865 & 31.374 \\
\texttt{li-test4-94} & 36809 & 142491 & \textbf{141863} & 2.556 & 141843 & 29.103 \\
\texttt{li-test4-95} & 36809 & 142491 & 141850 & 58.823 & 141868 & 16.500 \\
\texttt{li-test4-96} & 36809 & 142491 & \textbf{141858} & 6.038 & 141848 & 16.622 \\
\texttt{li-test4-97} & 36809 & 142491 & \textbf{141855} & 29.113 & 141852 & 33.242 \\
\texttt{li-test4-98} & 36809 & 142491 & \textbf{141862} & 40.632 & 141850 & 52.114 \\
\texttt{li-test4-99} & 36809 & 142491 & \textbf{141859} & 12.537 & 141855 & 31.298 \\
\texttt{gripper10u} & 2312 & 18666 & \textbf{18663} & \textbf{0.877} & 18663 & 2.014 \\
\texttt{gripper11u} & 3084 & 26019 & \textbf{26017} & 15.150 & 26016 & 6.459 \\
\texttt{gripper12u} & 3352 & 29412 & \textbf{29409} & \textbf{14.485} & 29409 & 29.184 \\
\texttt{gripper13u} & 4268 & 38965 & 38961 & 9.830 & 38961 & 1.315 \\
\texttt{gripper14u} & 4584 & 43390 & \textbf{43386} & \textbf{23.381} & 43386 & 36.584 \\
\texttt{bc56-sensors-1-k391-unsat} & 561371 & 1778987 & \textbf{1600252} & 15.997 & 1600079 & 43.783 \\
\texttt{bc56-sensors-2-k592-unsat} & 850398 & 2694319 & 2366874 & 48.252 & 2366923 & 13.745 \\
\texttt{bc57-sensors-1-k303-unsat} & 435701 & 1379987 & 1261961 & 39.150 & 1262043 & 14.386 \\
\texttt{dme-03-1-k247-unsat} & 261352 & 773077 & \textbf{736229} & 49.430 & 736010 & 22.759 \\
\texttt{motors-stuck-1-k407-unsat} & 654766 & 2068742 & \textbf{1842393} & 55.420 & 1842299 & 7.294 \\
\texttt{motors-stuck-2-k314-unsat} & 505536 & 1596837 & 1445274 & 53.360 & 1445323 & 57.764 \\
\texttt{valves-gates-1-k617-unsat} & 985042 & 3113540 & 2714448 & 58.357 & 2714681 & 30.268 \\
\texttt{6pipe} & 15800 & 394739 & \textbf{394717} & 42.251 & - & - \\
\texttt{7pipe} & 23910 & 751118 & - & - & - & - \\
\texttt{comb1} & 5910 & 16804 & 16749 & 43.330 & 16751 & 44.024 \\
\texttt{dp12u11} & 11137 & 30792 & 30785 & 11.240 & 30789 & 17.881 \\
\texttt{f2clk\_50} & 34678 & 101319 & 100629 & 18.230 & 100668 & 29.506 \\
\texttt{fifo8\_300} & 194762 & 530713 & 506270 & 9.520 & 506329 & 49.814 \\
\texttt{homer17} & 286 & 1742 & \textbf{1738} & \textbf{0.126} & 1738 & 0.317 \\
\texttt{homer18} & 308 & 2030 & \textbf{2024} & \textbf{0.131} & 2024 & 0.331 \\
\texttt{homer19} & 330 & 2340 & \textbf{2332} & \textbf{0.142} & 2332 & 0.349 \\
\texttt{homer20} & 440 & 4220 & \textbf{4202} & \textbf{0.170} & 4202 & 0.429 \\
\texttt{k2fix\_gr\_2pinvar\_w8} & 3771 & 270136 & \textbf{269918} & 47.860 & 269910 & 29.096 \\
\texttt{k2fix\_gr\_2pinvar\_w9} & 5028 & 307674 & 307563 & 43.461 & 307565 & 42.008 \\
\texttt{k2fix\_gr\_2pin\_w8} & 9882 & 295998 & 295657 & 54.870 & 295691 & 5.697 \\
\texttt{k2fix\_gr\_2pin\_w9} & 13176 & 345426 & \textbf{345230} & 39.450 & 345228 & 24.266 \\
\texttt{k2fix\_gr\_rcs\_w8} & 10056 & 271393 & \textbf{271296} & 17.291 & 271292 & 58.947 \\
\texttt{sha1} & 61377 & 255417 & 251863 & 54.40 & 251927 & 48.348 \\
\texttt{sha2} & 61377 & 255417 & \textbf{251915} & 19.47 & 251873 & 17.809 \\ \hline
\texttt{rsdecoder1\_blackbox\_KESblock} & 707330 & 1106376 & \textbf{1030025} & 5.471 & 1030001 & 47.969 \\
\texttt{rsdecoder4.dimacs} & 237783 & 933978 & 896327 & 16.880 & 896441 & 51.221 \\
\texttt{rsdecoder-problem.dimacs\_38} & 1198012 & 3865513 & 3350748 & 11.820 & 3351073 & 46.902 \\
\texttt{rsdecoder-problem.dimacs\_41} & 1186710 & 3829036 & 3320154 & 37.064 & 3320274 & 36.249 \\
\texttt{SM\_MAIN\_MEM\_buggy1.dimacs} & 870975 & 3812147 & 3416609 & 43.833 & 3416837 & 37.047 \\
\texttt{wb\_4m8s1.dimacs} & 463080 & 1759150 & \textbf{1624751} & 17.008 & 1624325 & 27.448 \\
\texttt{wb\_4m8s4.dimacs} & 463080 & 1759150 & \textbf{1624017} & 4.333 & 1623800 & 22.801 \\
\texttt{wb\_4m8s-problem.dimacs\_47} & 2691648 & 8517027 & \textbf{7159756} & 56.460 & 7159458 & 53.487 \\
\texttt{wb\_4m8s-problem.dimacs\_49} & 2785108 & 8812799 & 7401876 & 43.930 & 7402080 & 52.858 \\
\texttt{wb\_conmax1.dimacs} & 277950 & 1221020 & \textbf{1168273} & 23.774 & 1168267 & 27.155 \\
\texttt{wb\_conmax3.dimacs} & 277950 & 1221020 & \textbf{1168336} & 1.808 & 1168173 & 56.621 \\ \hline
\end{tabular}

\label{nov}
\end{table}

\begin{table}[p]
\centering
\scriptsize
\caption{Results for $H=\textit{walksat-tabu}$. Times are given in minutes.}
\begin{tabular}{lcccrcr}
\hline & & & \multicolumn{2}{c}{Genetic algorithm} & \multicolumn{2}{c}{\textit{walksat-tabu} alone} \\
Instance & $n$ & $m$ & {Num.\ sat.\ cl.} & Time & {Num.\ sat.\ cl.} & Time \\ \hline
\texttt{c3540mul} &5248 & 33199 & 33164 & 24.820 & 33166 & 59.816 \\
\texttt{c6288mul} &9540 & 61421 & \textbf{61392} & 3.884 & 61389 & 23.451 \\
\texttt{dalumul} &9426 & 59991 & 59896 & 2.730 & 59910 & 2.546 \\
\texttt{frg1mul} & 3230 & 20575 & 20570 & 30.706 & 20570 & 6.465 \\
\texttt{k2mul} & 11680 & 74581 & 74340 & 6.720 & 74341 & 21.223 \\
\texttt{x1mul} & 8760 & 55571 & 55561 & 34.556 & 55562 & 26.049 \\
\texttt{am\_6\_6} & 2269 & 7814 & 7809 & 2.580 & 7810 & 58.420 \\
\texttt{am\_7\_7} & 4264 & 14751 & \textbf{14732} & 29.810 & 14729 & 0.926 \\
\texttt{am\_8\_8} & 7361 & 25538 & 25466 & 19.600 & 25473 & 45.678 \\
\texttt{am\_9\_9} & 11908 & 41393 & 41183 & 34.960 & 41189 & 59.298 \\
\texttt{li-exam-61} & 28147 & 108436 & \textbf{107983} & 14.918 & 107982 & 31.107 \\
\texttt{li-exam-62} & 28147 & 108436 & 107974 & 41.560 & 107992 & 53.727 \\
\texttt{li-exam-63} & 28147 & 108436 & \textbf{107980} & 48.127 & 107978 & 58.207 \\
\texttt{li-exam-64} & 28147 & 108436 & \textbf{107985} & \textbf{3.946} & 107985 & 20.522 \\
\texttt{li-test4-100} & 36809 & 142491 & \textbf{141790} & 39.720 & 141782 & 21.592 \\
\texttt{li-test4-101} & 36809 & 142491 & \textbf{141804} & 19.040 & 141800 & 11.141 \\
\texttt{li-test4-94} & 36809 & 142491 & \textbf{141806} & 2.853 & 141780 & 44.018 \\
\texttt{li-test4-95} & 36809 & 142491 & \textbf{141791} & 35.540 & 141782 & 27.593 \\
\texttt{li-test4-96} & 36809 & 142491 & \textbf{141804} & 24.995 & 141791 & 33.641 \\
\texttt{li-test4-97} & 36809 & 142491 & 141791 & 26.641 & 141805 & 22.784 \\
\texttt{li-test4-98} & 36809 & 142491 & 141777 & 26.300 & 141791 & 54.528 \\
\texttt{li-test4-99} & 36809 & 142491 & 141774 & 37.930 & 141781 & 51.666 \\
\texttt{gripper10u} & 2312 & 18666 & \textbf{18662} & \textbf{42.287} & 18662 & 58.256 \\
\texttt{gripper11u} & 3084 & 26019 & \textbf{26014} & \textbf{2.383} & 26014 & 10.450 \\
\texttt{gripper12u} & 3352 & 29412 & \textbf{29406} & \textbf{2.104} & 29406 & 2.788 \\
\texttt{gripper13u} & 4268 & 38965 & \textbf{38959} & \textbf{2.370} & 38959 & 50.868 \\
\texttt{gripper14u} & 4584 & 43390 & \textbf{43383} & 2.520 & 43382 & 0.512 \\
\texttt{bc56-sensors-1-k391-unsat} & 561371 & 1778987 & \textbf{1600245} & 36.329 & 1599963 & 43.683 \\
\texttt{bc56-sensors-2-k592-unsat} & 850398 & 2694319 & \textbf{2361372} & 33.625 & 2361225 & 39.390 \\
\texttt{bc57-sensors-1-k303-unsat} & 435701 & 1379987 & \textbf{1264444} & 33.863 & 1264039 & 44.873 \\
\texttt{dme-03-1-k247-unsat} & 261352 & 773077 & \textbf{740068} & 15.573 & 739924 & 53.964 \\
\texttt{motors-stuck-1-k407-unsat} & 654766 & 2068742 & \textbf{1840274} & 36.547 & 1839998 & 36.970 \\
\texttt{motors-stuck-2-k314-unsat} & 505536 & 1596837 & \textbf{1445995} & 41.870 & 1445813 & 34.788 \\
\texttt{valves-gates-1-k617-unsat} & 985042 & 3113540 & \textbf{2705763} & 55.140 & 2705761 & 53.865 \\
\texttt{6pipe} & 15800 & 394739 & \textbf{394727} & 48.824 & - & - \\
\texttt{7pipe} & 23910 & 751118 & \textbf{751102} & 27.336 & - & - \\
\texttt{comb1} & 5910 & 16804 & 16713 & 59.406 & 16717 & 9.986 \\
\texttt{dp12u11} & 11137 & 30792 & \textbf{30775} & 22.338 & 30773 & 27.777 \\
\texttt{f2clk\_50} & 34678 & 101319 & \textbf{100087} & 49.690 & 100075 & 27.279 \\
\texttt{fifo8\_300} & 194762 & 530713 & 509196 & 2.030 & 509252 & 56.291 \\
\texttt{homer17} & 286 & 1742 & \textbf{1738} & \textbf{0.105} & 1738 & 0.268 \\
\texttt{homer18} & 308 & 2030 & \textbf{2024} & \textbf{0.112} & 2024 & 0.285 \\
\texttt{homer19} & 330 & 2340 & \textbf{2332} & \textbf{0.119} & 2332 & 0.309 \\
\texttt{homer20} & 440 & 4220 & \textbf{4202} & \textbf{0.152} & 4202 & 0.374 \\
\texttt{k2fix\_gr\_2pinvar\_w8} & 3771 & 270136 & 269839 & 46.921 & 269855 & 34.207 \\
\texttt{k2fix\_gr\_2pinvar\_w9} & 5028 & 307674 & \textbf{307490} & 31.670 & 307485 & 14.594 \\
\texttt{k2fix\_gr\_2pin\_w8} & 9882 & 295998 & 295544 & 12.160 & 295554 & 43.311 \\
\texttt{k2fix\_gr\_2pin\_w9} & 13176 & 345426 & \textbf{345065} & 3.299 & 345044 & 47.063 \\
\texttt{k2fix\_gr\_rcs\_w8} & 10056 & 271393 & \textbf{271301} & \textbf{42.525} & 271301 & 48.211 \\
\texttt{sha1} & 61377 & 255417 & \textbf{251374} & 30.88 & 251364 & 9.204 \\
\texttt{sha2} & 61377 & 255417 & 251390 & 7.67 & 251392 & 16.520 \\ \hline
\texttt{rsdecoder1\_blackbox\_KESblock} & 707330 & 1106376 & 1028280 & 55.013 & 1028328 & 33.828 \\
\texttt{rsdecoder4.dimacs} & 237783 & 933978 & 895865 & 1.950 & 895990 & 16.142 \\
\texttt{rsdecoder-problem.dimacs\_38} & 1198012 & 3865513 & 3334250 & 34.230 & 3334450 & 49.290 \\
\texttt{rsdecoder-problem.dimacs\_41} & 1186710 & 3829036 & 3304293 & 7.626 & 3304476 & 4.832 \\
\texttt{SM\_MAIN\_MEM\_buggy1.dimacs} & 870975 & 3812147 & 3386126 & 37.130 & 3386263 & 58.493 \\
\texttt{wb\_4m8s1.dimacs} & 463080 & 1759150 & 1611716 & 56.910 & 1611839 & 59.405 \\
\texttt{wb\_4m8s4.dimacs} & 463080 & 1759150 & 1610945 & 36.070 & 1611361 & 56.378 \\
\texttt{wb\_4m8s-problem.dimacs\_47} & 2691648 & 8517027 & 7133234 & 33.360 & 7133580 & 44.962 \\
\texttt{wb\_4m8s-problem.dimacs\_49} & 2785108 & 8812799 & 7374846 & 3.900 & 7374877 & 32.896 \\
\texttt{wb\_conmax1.dimacs} & 277950 & 1221020 & 1156736 & 35.785 & 1156771 & 22.056 \\
\texttt{wb\_conmax3.dimacs} & 277950 & 1221020 & 1156718 & 31.800 & 1156841 & 35.277 \\ \hline
\end{tabular}

\label{wt}
\end{table}

\begin{table}[p]
\centering
\scriptsize
\caption{Results for $H=\textit{adaptnovelty+}$. Times are given in minutes.}
\begin{tabular}{lcccrcr}
\hline & & & \multicolumn{2}{c}{Genetic algorithm} & \multicolumn{2}{c}{\textit{adaptnovelty+} alone} \\
Instance & $n$ & $m$ & {Num.\ sat.\ cl.} & Time & {Num.\ sat.\ cl.} & Time \\ \hline
\texttt{c3540mul} &5248 & 33199 & \textbf{33162} & 52.030 & 33153 & 30.089 \\
\texttt{c6288mul} &9540 & 61421 & \textbf{61382} & \textbf{1.893} & 61382 & 36.361 \\
\texttt{dalumul} &9426 & 59991 & \textbf{59930} & 47.250 & 59920 & 44.117 \\
\texttt{frg1mul} & 3230 & 20575 & \textbf{20574} & \textbf{0.527} & 20574 & 1.290 \\
\texttt{k2mul} & 11680 & 74581 & \textbf{74417} & 52.360 & 74407 & 16.932 \\
\texttt{x1mul} & 8760 & 55571 & \textbf{55570} & \textbf{1.153} & 55570 & 2.477 \\
\texttt{am\_6\_6} & 2269 & 7814 & \textbf{7813} & \textbf{0.102} & 7813 & 0.267 \\
\texttt{am\_7\_7} & 4264 & 14751 & \textbf{14750} & \textbf{0.277} & 14750 & 0.558 \\
\texttt{am\_8\_8} & 7361 & 25538 & \textbf{25523} & 44.450 & 25520 & 8.450 \\
\texttt{am\_9\_9} & 11908 & 41393 & \textbf{41296} & 49.254 & 41293 & 53.522 \\
\texttt{li-exam-61} & 28147 & 108436 & \textbf{108037} & 23.833 & 108028 & 54.671 \\
\texttt{li-exam-62} & 28147 & 108436 & \textbf{108045} & 38.670 & 108032 & 17.074 \\
\texttt{li-exam-63} & 28147 & 108436 & \textbf{108031} & 44.030 & 108026 & 49.837 \\
\texttt{li-exam-64} & 28147 & 108436 & 108030 & 41.830 & 108037 & 49.076 \\
\texttt{li-test4-100} & 36809 & 142491 & \textbf{141903} & 16.011 & 141902 & 35.010 \\
\texttt{li-test4-101} & 36809 & 142491 & 141894 & 4.150 & 141897 & 15.527 \\
\texttt{li-test4-94} & 36809 & 142491 & 141899 & 38.595 & 141908 & 37.992 \\
\texttt{li-test4-95} & 36809 & 142491 & \textbf{141900} & 17.200 & 141890 & 57.627 \\
\texttt{li-test4-96} & 36809 & 142491 & \textbf{141902} & 31.204 & 141896 & 52.649 \\
\texttt{li-test4-97} & 36809 & 142491 & 141897 & 27.190 & 141920 & 33.317 \\
\texttt{li-test4-98} & 36809 & 142491 & 141898 & 34.560 & 141907 & 52.417 \\
\texttt{li-test4-99} & 36809 & 142491 & 141902 & 29.421 & 141906 & 12.445 \\
\texttt{gripper10u} & 2312 & 18666 & \textbf{18665} & \textbf{0.161} & 18665 & 5.913 \\
\texttt{gripper11u} & 3084 & 26019 & 26018 & 18.106 & 26018 & 15.914 \\
\texttt{gripper12u} & 3352 & 29412 & 29410 & 0.600 & 29411 & 56.930 \\
\texttt{gripper13u} & 4268 & 38965 & 38963 & 4.270 & 38963 & 0.454 \\
\texttt{gripper14u} & 4584 & 43390 & 43388 & 23.785 & 43388 & 0.456 \\
\texttt{bc56-sensors-1-k391-unsat} & 561371 & 1778987 & 1623357 & 17.990 & 1623430 & 36.854 \\
\texttt{bc56-sensors-2-k592-unsat} & 850398 & 2694319 & \textbf{2394308} & 40.210 & 2393894 & 6.102 \\
\texttt{bc57-sensors-1-k303-unsat} & 435701 & 1379987 & 1282208 & 35.553 & 1282338 & 34.813 \\
\texttt{dme-03-1-k247-unsat} & 261352 & 773077 & 746902 & 43.755 & 746935 & 19.452 \\
\texttt{motors-stuck-1-k407-unsat} & 654766 & 2068742 & 1866990 & 12.900 & 1867266 & 56.079 \\
\texttt{motors-stuck-2-k314-unsat} & 505536 & 1596837 & \textbf{1467201} & 55.320 & 1467123 & 43.089 \\
\texttt{valves-gates-1-k617-unsat} & 985042 & 3113540 & \textbf{2742641} & 48.170 & 2742310 & 18.078 \\
\texttt{6pipe} & 15800 & 394739 & \textbf{393808} & 46.970 & 393771 & 41.444 \\
\texttt{7pipe} & 23910 & 751118 & \textbf{749636} & 54.051 & - & - \\
\texttt{comb1} & 5910 & 16804 & 16756 & 29.750 & 16759 & 31.832 \\
\texttt{dp12u11} & 11137 & 30792 & \textbf{30723} & 15.480 & 30722 & 12.257 \\
\texttt{f2clk\_50} & 34678 & 101319 & 100431 & 6.983 & 100435 & 3.063 \\
\texttt{fifo8\_300} & 194762 & 530713 & 516316 & 55.551 & 516380 & 35.441 \\
\texttt{homer17} & 286 & 1742 & \textbf{1738} & \textbf{0.133} & 1738 & 0.330 \\
\texttt{homer18} & 308 & 2030 & \textbf{2024} & \textbf{0.138} & 2024 & 0.345 \\
\texttt{homer19} & 330 & 2340 & \textbf{2332} & \textbf{0.143} & 2332 & 0.367 \\
\texttt{homer20} & 440 & 4220 & \textbf{4202} & \textbf{0.175} & 4202 & 0.440 \\
\texttt{k2fix\_gr\_2pinvar\_w8} & 3771 & 270136 & 269923 & 30.530 & 269930 & 39.335 \\
\texttt{k2fix\_gr\_2pinvar\_w9} & 5028 & 307674 & \textbf{307560} & 47.330 & 307559 & 28.284 \\
\texttt{k2fix\_gr\_2pin\_w8} & 9882 & 295998 & \textbf{295646} & 55.630 & 295645 & 11.987 \\
\texttt{k2fix\_gr\_2pin\_w9} & 13176 & 345426 & 345123 & 38.150 & 345150 & 51.661 \\
\texttt{k2fix\_gr\_rcs\_w8} & 10056 & 271393 & 271282 & 25.350 & 271290 & 39.640 \\
\texttt{sha1} & 61377 & 255417 & \textbf{252622} & 34.02 & 252613 & 52.078 \\
\texttt{sha2} & 61377 & 255417 & 252615 & 34.58 & 252617 & 22.981 \\ \hline
\texttt{rsdecoder1\_blackbox\_KESblock} & 707330 & 1106376 & 1042736 & 58.190 & 1042849 & 35.251 \\
\texttt{rsdecoder4.dimacs} & 237783 & 933978 & 904931 & 50.378 & 905006 & 38.068 \\
\texttt{rsdecoder-problem.dimacs\_38} & 1198012 & 3865513 & \textbf{3374308} & 33.560 & 3374204 & 28.959 \\
\texttt{rsdecoder-problem.dimacs\_41} & 1186710 & 3829036 & \textbf{3343967} & 55.770 & 3343653 & 12.919 \\
\texttt{SM\_MAIN\_MEM\_buggy1.dimacs} & 870975 & 3812147 & \textbf{3431928} & 15.243 & 3431455 & 18.274 \\
\texttt{wb\_4m8s1.dimacs} & 463080 & 1759150 & \textbf{1637425} & 6.646 & 1637337 & 25.506 \\
\texttt{wb\_4m8s4.dimacs} & 463080 & 1759150 & \textbf{1636787} & 15.974 & 1636718 & 49.083 \\
\texttt{wb\_4m8s-problem.dimacs\_47} & 2691648 & 8517027 & 7189090 & 14.730 & 7189244 & 19.575 \\
\texttt{wb\_4m8s-problem.dimacs\_49} & 2785108 & 8812799 & \textbf{7432769} & 20.240 & 7432463 & 12.193 \\
\texttt{wb\_conmax1.dimacs} & 277950 & 1221020 & 1175783 & 43.730 & 1175804 & 8.269 \\
\texttt{wb\_conmax3.dimacs} & 277950 & 1221020 & 1175833 & 12.660 & 1175942 & 14.228 \\ \hline
\end{tabular}

\label{anov}
\end{table}

Some entries in the tables are highlighted by a bold typeface to indicate that
the genetic algorithm found a solution strictly better than the one found by the
heuristic when used by itself, or a solution satisfying the same number of
clauses but first encountered in a shorter time. In the former case only the
number of satisfied clauses is highlighted, in the latter case the time is
highlighted as well. The number of highlighted instances amounts to the
ratios given in Table~\ref{ratios}. Clearly, with the notable exception of
$H=\textit{walksat-tabu}$ on the 2008 dataset (on which the use of $H$ alone
outperformed the genetic algorithm on all $11$ instances), the genetic algorithm
succeeds well on a significant fraction of the instances.

\begin{table}
\centering
\caption{Success ratios of the genetic algorithm as per
Tables~\ref{nov} through \ref{anov}.}
\begin{tabular}{lccc}
\hline Instance set & \textit{novelty} & \textit{walksat-tabu} & \textit{adaptnovelty+} \\ \hline
2004 dataset & $0.540$ & $0.667$ & $0.588$ \\
2008 dataset & $0.545$ & $0.000$ & $0.545$ \\
2004 \& 2008 datasets combined & $0.541$ & $0.548$ & $0.581$ \\ \hline
\end{tabular}

\label{ratios}
\end{table}

Revising these ratios to contemplate all instances from both datasets (i.e.,
include the results omitted from Tables~\ref{nov} through~\ref{anov}) yields the
ratios in Table~\ref{ratiosall}. These show that the genetic algorithm fares
even better when evaluated on all $100$ instances of the 2004 dataset. They also
show slightly lower ratios for the genetic algorithm on the $112$-instance 2008
dataset for $H=\textit{novelty}$ and $H=\textit{adaptnovelty+}$. As for
$H=\textit{walksat-tabu}$, we see in Table~\ref{ratiosall} a dramatic increase
from the $0.000$ of Table~\ref{ratios}, indicating that for this particular $H$
on the complete 2008 dataset the genetic algorithm does better than the
heuristic alone only on the comparatively easier instances (and then for a
significant fraction of them).

\begin{table}
\centering
\caption{Success ratios of the genetic algorithm over all $100$ instances of the
2004 dataset and all $112$ instances of the 2008 dataset.}
\begin{tabular}{lccc}
\hline Instance set & \textit{novelty} & \textit{walksat-tabu} & \textit{adaptnovelty+} \\ \hline
2004 dataset & $0.600$ & $0.700$ & $0.630$ \\
2008 dataset & $0.518$ & $0.518$ & $0.509$ \\
2004 \& 2008 datasets combined & $0.557$ & $0.604$ & $0.566$ \\ \hline
\end{tabular}

\label{ratiosall}
\end{table}

\section{Concluding remarks}
\label{conclusions}

Given a heuristic for some problem of combinatorial optimization, a preamble
such as we defined in Section~\ref{preambles} for MAX-SAT is a selector of
initial conditions. As such, it aims at isolating the inevitable randomness of
the initial conditions one normally uses with such heuristics from the heuristic
itself. By doing so, preambles attempt to poise the heuristic to operate from
more favorable initial conditions.

In this paper we have demonstrated the success of MAX-SAT preambles when they
are discovered, given the MAX-SAT instance and heuristic of interest, via an
evolutionary algorithm. As we showed in Section~\ref{results}, for well
established benchmark instances and heuristics the resulting genetic algorithm
can outperform the heuristics themselves when used alone. We believe further
effort can be profitably spent on attempting similar solutions to other problems
that, like MAX-SAT, can be expressed as an unconstrained optimization problem on
binary variables. Some of them are the maximum independent set and minimum
dominating set problems on graphs, both admitting well-known formulations of
this type \cite{bg89}.

\subsection*{Acknowledgments}

The authors acknowledge partial support from CNPq, CAPES, and a FAPERJ BBP
grant.

\bibliography{maxsat}
\bibliographystyle{plain}

\end{document}